\pdfoutput=1
\documentclass[11pt]{article}

\usepackage{EMNLP2022}

\usepackage{times}
\usepackage{latexsym}

\usepackage[T1]{fontenc}
\usepackage[utf8]{inputenc}

\usepackage{microtype}
\usepackage{colortbl}
\usepackage{color}
\usepackage{microtype}
\usepackage{mathtools}
\usepackage{paralist}
\usepackage{subcaption}
\usepackage{multirow}
\usepackage{booktabs}
\usepackage{graphicx}
\usepackage{tabularx}
\usepackage{cleveref}
\usepackage{caption}
\usepackage{adjustbox}
\usepackage{xspace}
\usepackage{arydshln}
\crefformat{section}{\S#2#1#3}
\crefformat{subsection}{\S#2#1#3}
\crefformat{subsubsection}{\S#2#1#3}
\newcolumntype{Z}{>{\raggedleft\arraybackslash}X}
\newcolumntype{Y}{>{\centering\arraybackslash}X}
\graphicspath{ {./images/} }
\definecolor{dark_green}{rgb}{0.0, 0.6, 0.0}
\definecolor{light_gray}{rgb}{0.9, 0.9, 0.9}
\counterwithin*{equation}{section}

\usepackage{enumitem}
\setlist[enumerate]{itemsep=0mm}

\newcommand{\dataset}{MVQG\xspace}
\title{Multi-VQG: Generating Engaging Questions for Multiple Images}

\author{Min-Hsuan Yeh$^\ast$, Vicent Chen$^\diamond$, Ting-Hao `Kenneth' Haung$^\dagger$, Lun-Wei Ku$^\ddagger$\\
  University of Massachusetts Amherst$^\ast$,  University of Illinois Urbana-Champaign$^\diamond$,\\ 
  Pennsylvania State University$^\dagger$, 
  Institute of Information Science, Academia Sinica$^\ddagger$ \\
  \texttt{myeh@umass.edu}, \texttt{vfchen2@illinois.edu},\\ \texttt{txh710@psu.edu}, \texttt{lwku@iis.sinica.edu}}

\begin{document}
\maketitle
\begin{abstract}

Generating engaging content has drawn much recent attention in the NLP community.
Asking questions is a natural way to respond to photos and promote
awareness.
However, most answers to questions in traditional
question-answering (QA) datasets are factoids, which reduce individuals'
willingness to answer. Furthermore, traditional visual question generation
(VQG) confines the source data for question generation to single images,
resulting in a limited ability to comprehend time-series information of the underlying
event. In this paper, we propose generating engaging
questions from multiple images. We present \dataset\footnote{Github repo:
\href{https://github.com/AcademiaSinicaNLPLab/MVQG-Dataset-of-Generating-Engaging-Questions-for-Multiple-Images}{https://github.com/AcademiaSinicaNLPLab /MVQG-Dataset-of-Generating-Engaging-Questions-for-Mu ltiple-Images}}, a new dataset, and establish
a series of baselines, including both end-to-end and dual-stage architectures.
Results show that building stories behind the image sequence enables models to
generate engaging questions, which confirms our assumption that people
typically construct a picture of the event in their minds before asking
questions. These results open up an exciting challenge for visual-and-language
models to implicitly construct a story behind a series of photos to allow for
creativity and experience sharing and hence draw attention to downstream
applications.

%According to data analysis, our instructions aid in the collection of more engaging questions than VQG, the prior visual question generation dataset. 

\end{abstract}

\section{Introduction}

The popularity of image-sharing behavior in chats and social media applications
shows that this is a natural way to increase participant
engagement~\citep{Hu2014WhatWI}. In response, asking questions based on these
photos is a straightforward method to promote awareness, sustain attention, and
acquire useful information. An obvious example is that when we see someone
share a photo of a car accident on Facebook, commenting ``Was anyone injured in
the crash?''
draws more attention and
replies from both the author and other readers than ``Oh my goodness, that is
serious.'' Furthermore, from the author's aspect, posting photos on social
media with an engaging, image-related question helps the author to hear the
public voice of their feelings and thoughts, and keeps the author connected
with the world~\citep{lu-etal-2021-engage}. 

\begin{figure}[t]
     \centering
     \includegraphics[width=\linewidth]{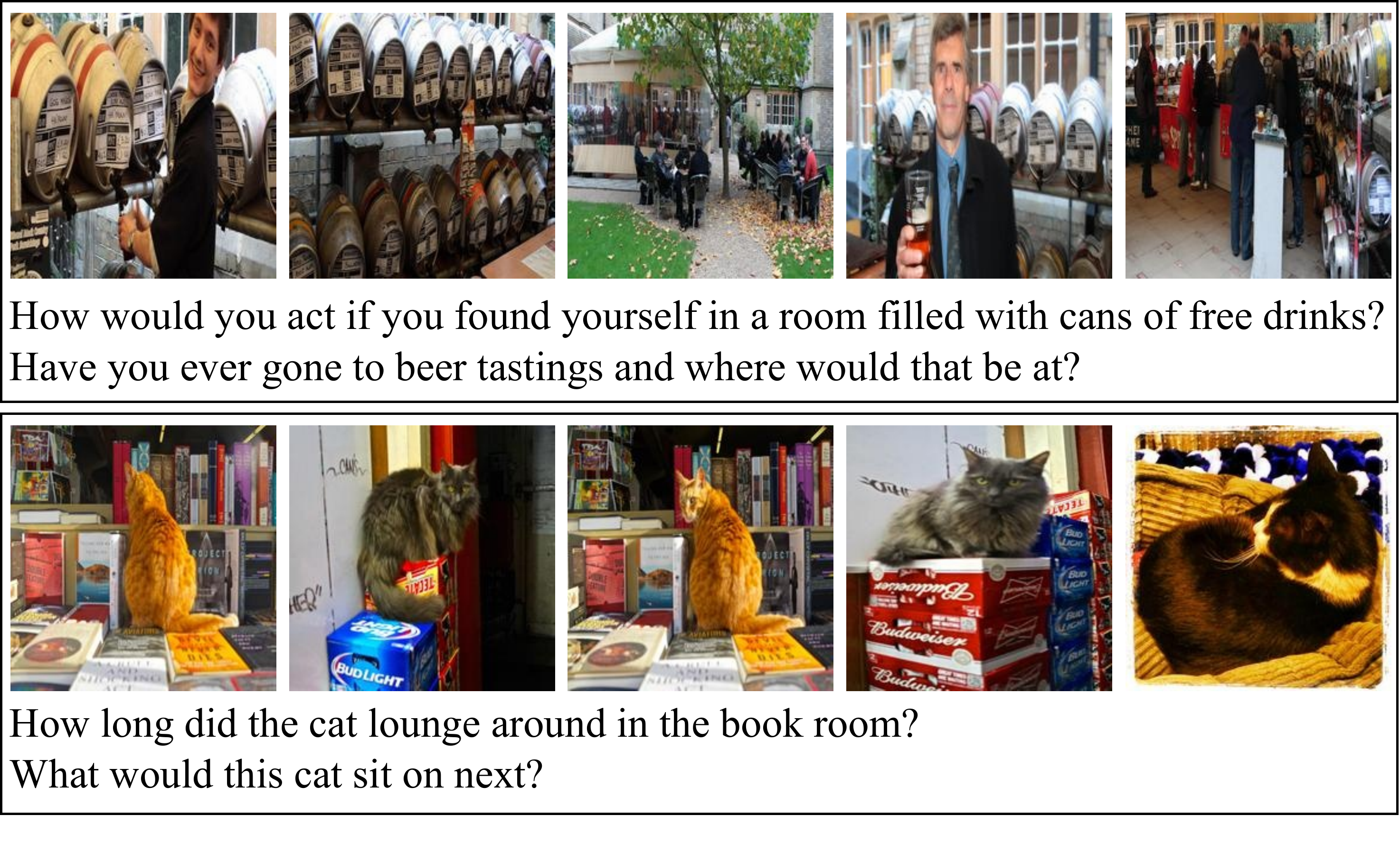}
% 	  \vspace{-1.8pc}
	  \caption{Two examples of \dataset. Each data point consists of an image sequence and two to five engaging questions written by humans. The machine should generate a question over a given image sequence.
	  }
% 	  \vspace{-.5pc}
\label{fig:example}
\end{figure}
 
However, not all the questions have the same effect. Questions in 
traditional text-based QA datasets such as
SQuAD~\citep{rajpurkar-etal-2016-squad}, NarrativeQA~\citep{Kocisk2018TheNR},
or FairytaleQA~\citep{zhao-etal-2022-educational} are for educational purposes
and language understanding, which do not seek to encourage people to reply.
Meanwhile,
questions in the VQA dataset~\citep{VQA} usually ask about the
color of an object or its position, to which the answers are too obvious for
humans to respond. In fact, these two kinds of questions are rarely seen in 
daily chat and social media posts. \citet{Shuster2019EngagingIC} state that
humans consider engaging and effective captions those that ``avoid
stating the obvious.'' As with image captions, an engaging and effective
question asks about things behind the scenes and is usually open-ended.

Moreover, the input information for question generation also matters. Images
are more straightforward for humans than text, and they provide plenty of room
for imagination. In addition, images shared on social media are often
sequential instead of solitary, as a single image gives readers only a limited
understanding of the experience being shared. To this end, despite the existence
of question generation (QG) datasets such as those created by
\citet{lu-etal-2021-engage}, \citet{Wei2021VisualQR}, or
\citet{DBLP:journals/corr/MostafazadehMDZ16}, which contain engaging questions,
their text-based or single-image settings limit the usage of current QG models
in popular applications. 

To facilitate downstream applications involving a set of shared images, e.g.,
accompanying robots, social media robots, automatic assistants, or reminiscence
therapy, we propose generating an engaging question from multiple images. We
create \dataset, a new dataset, by asking workers to write down a question
following instructions based on a sequence of photos from
VIST~\cite{huang2016visual}, a dataset consisting of five sequential
images and a story about those images. For a better illustration of the task,
Figure~\ref{fig:example} shows two examples of \dataset. 
% \mh{
Unlike the instruction of VQA's data collection~\citep{Antol_2015_ICCV} asked workers to imagine ``a smart robot`` that ``understands a lot about images,`` such as objects, scenes, or color, or texture, and come up with questions to ``stump this smart robot.'' Our instruction, on the other hand, asked workers to imagine that they want to have a conversation with people on Twitter and hence to write a question to start that conversation.
% }
% The goal of the proposed task is to generate an engaging question given an image sequence.
The data analysis shows that our instructions help collect more engaging
questions than VQG~\citep{DBLP:journals/corr/MostafazadehMDZ16}, the benchmark
dataset for visual question generation. Furthermore, we establish a series of
baselines, including both end-to-end and dual-stage architectures. The
experimental results show that information about stories behind the image
sequence helps baselines generate engaging questions, which confirms our
assumption that humans typically construct stories in their heads before asking
questions. These results open up an exciting challenge for visual-and-language
models: implicitly constructing a story behind a series of photos to allow
for creativity and experience sharing, hence drawing attention to its
downstream applications.

The contributions of our paper are threefold: first, we introduce 
a novel task multi-VQG and \dataset, a new dataset:  
given a sequence of relevant images,
generate a corresponding engaging question; second, we propose several
baselines and show that story information helps baselines to generate engaging
questions from image sequences; third, we propose five aspects for human
evaluation as benchmarks to better evaluate the engagement of generated
questions.

\section{Related Work}

User engagement has received much recent attention in the NLP community.
\citet{DBLP:journals/corr/MostafazadehMDZ16} created the first visual
question generation dataset comprised of natural and engaging questions.
However, engagement is not well-stated in this work; it simply means
``the first question that comes to mind''.
\citet{Shuster2019EngagingIC} present an engaging image captioning task to
improve the ability of machines to communicate with humans; engaging
captions are defined as captions that ``avoid stating the obvious.''
\citet{lu-etal-2021-engage} develop a dataset for poll-question generation
for social media posts. This work demonstrates that the poll question is an
engaging question that can be utilized to help us hear the public voice
for decision-making and thus better understand our society.
\citet{Wei2021VisualQR} state that an engaging and attractive question may
incorporate additional details or emotional phrases. Such questions are
more likely to be answered. 
Images and questions are two prominent elements in these works, indicating that
visual stimulation and inquiry are typical means to communicate awareness and
sustain connections. However, these studies primarily consider 
single images, limiting the use of current QG models in popular
applications because individuals typically share multiple photos to express
more comprehensive experiences. In our study, we propose generating engaging
questions over an image sequence and creating a dataset comprised of five
photos and human-written questions.

A visual-and-language (VL) model is typically used to generate engaging
questions from images. After the development of
BERT~\citep{devlin-etal-2019-bert}, various BERT-based VL models were
proposed. These VL models are designed to integrate information from both
vision and language modalities via an encoder, and are categorized into
fusion encoders and dual encoders based on how input from
distinct modalities is
aggregated~\citep{https://doi.org/10.48550/arxiv.2202.10936}. Fusion encoder
models such as VisualBERT~\citep{https://doi.org/10.48550/arxiv.1908.03557},
XLMERT~\citep{cho-etal-2020-x}, SOHO~\citep{Huang2021SeeingOO}, and
VL-T5~\citep{cho2021vlt5} encode text embeddings and image features in
the same model with different fusion approaches. Following self- or
cross-attention, the hidden state of the last layer is treated as a fused
representation of different modalities. Because fusion encoder models require
image and text pairings as input, the model must input all possible pairs in
image-text matching tasks, resulting in poor inference speed. Dual encoder
models such as CLIP~\citep{Radford2021LearningTV}, on the other hand, use two
single-modal encoders to encode the two modalities separately and use the dot
product to project the image embedding and text embedding to the same semantic
space to compute VL similarity scores. Although dual encoder models are
lighter, they frequently fail in difficult VL understanding tasks. As a result,
we continue to employ fusion encoder models as baselines in our work. We choose
VL-T5~\citep{cho2021vlt5} as the backbone in particular because it treats all
VL tasks as text-generating tasks, which is appropriate for our question
generation scenario. Inspired by \citet{shen2022how}, we propose an additional
baseline model by employing the visual encoder of
CLIP~\citep{Radford2021LearningTV} instead of the self-trained image feature
extractor in our fusion encoder, so that the image features are better
projected into the semantic space.

\section{Dataset: MVQG}

\subsection{Selection of Image Sequences}

Given that the proposed task is to generate engaging questions based on a
cohesive narrative, 
the input photographs cannot be
randomly selected from an image set. As a result, we choose image sequences
from the VIST dataset, the first dataset of sequential photos accompanied by
stories. 
% \mh{
In the VIST dataset, each image sequence containing five photos extracted from a Flickr album of a human event (e.g., ``wedding'' or ``first day of school''); five photos must be taken within a 48-hour span.
Workers constructing VIST arranged the five photos in the order chosen, and then wrote a
sentence for each photo to create a story.
% }
This procedure guaranteed that the
chosen image sequences were ``storyable'', i.e., they contained at least one
continuous narrative of an event or scene for question generation. 
% \mh{
In addition, although many social-media posts include multiple images that are not necessarily sequential, when social media users create a post that includes multiple photos, these photos often capture the same scene, event, or concept; that is, these photos can have similar properties to those in the VIST dataset.
% }
To this end,
we randomly chose 7,700 image sequences from the VIST training set and chose all 1,999
sequences from the VIST test set and assigned them to workers to annotate the
engaging questions.

\subsection{Question Annotations}

Human brains are excellent at object recognition; they can quickly recognize
the most significant details in photographs. However, finding the
relationship between visuals and developing a unified narrative of events or
scenes behind those items requires more time for humans. Thus, if 
workers are asked to write down a question immediately after seeing the image sequence,
they may merely inquire about the first object that comes to mind, rather than
ask engaging questions based on a cohesive narrative behind the photos. To
solve this problem, we created a data annotation approach to assist workers in
writing suitable sentences by answering a three-stage question:

\begin{enumerate}[nosep,label=Q\arabic*.]
	 \item Please list the top five objects (e.g., dogs, trees) or events (e.g.,
	 weddings, parties) you regard as being the most important in the image
	 sequence.
	 \item Please describe the visual sequence using one or more sentences based
	 on the items and events you observed in Q1.
	 \item Imagining that you decide to post this image sequence on Twitter and
	 want to expand the conversation by solely commenting on a question
	 connected to these images. What is the question you would ask based on the
	 description you gave in Q2?
\end{enumerate}

This strategy implicitly prompted workers to formulate an abstract notion of the
image sequence according to their observations. As a result, we were able to
obtain engaging questions that corresponded to the cohesive narratives of the
events depicted in the visual sequences. Furthermore, the descriptions provided
in Q2 qualify this dataset for multi-image captioning, making it suited for use
in a wider range of vision-and-language applications, e.g., image captioning.
Moreover, many recent question generation models are
answer-agnostic~\citep{https://doi.org/10.48550/arxiv.2203.08685,
https://doi.org/10.48550/arxiv.2203.04464}. Their findings show that adding
context summaries as the intermediary layer can improve the relevance and
interpretability of generated questions. Inspired by their research, the
descriptions provided in Q2 can also serve as summaries to generate
event-centric questions.

We gathered \dataset questions by crowdsourcing the task on Amazon Mechanical Turk (AMT). 
For each image sequence, we assigned 2~to~5 workers to annotate questions (at
\$0.2/HIT). We only accepted participants with a 98\% or more HIT acceptance
rate, had 3,000 or more finished HITs, and were located in the US. We also
required turkers to spend at least 30 seconds on each assignment. In total, we
asked workers to annotate 9,699 image sequences and obtained 31,421 questions.
After the annotation process, we manually revised the grammatical errors in
all questions. The dataset will be released after the paper is accepted.

\section{Dataset Analyses}

\subsection{Data Statistics}

\begin{table}[t]
    \centering
    \small
    \begin{tabular}{lc}
        \toprule
        \# all image sequences & 9,699 \\
        \# all questions & 31,421 \\
        \# all workers participated & 878\\
        Max \# questions written by one worker & 1,249\\
        Avg. \# questions written by one worker & 35.8\\
        \bottomrule
    \end{tabular}
    % \vspace{-0.5pc}
    \caption{Statistics of the annotation task.}
    % \vspace{-1.5pc}
    \label{tb:dataset_statistics}
\end{table}

\begin{figure}[t!]
    \centering
    \includegraphics[width=0.48\textwidth]{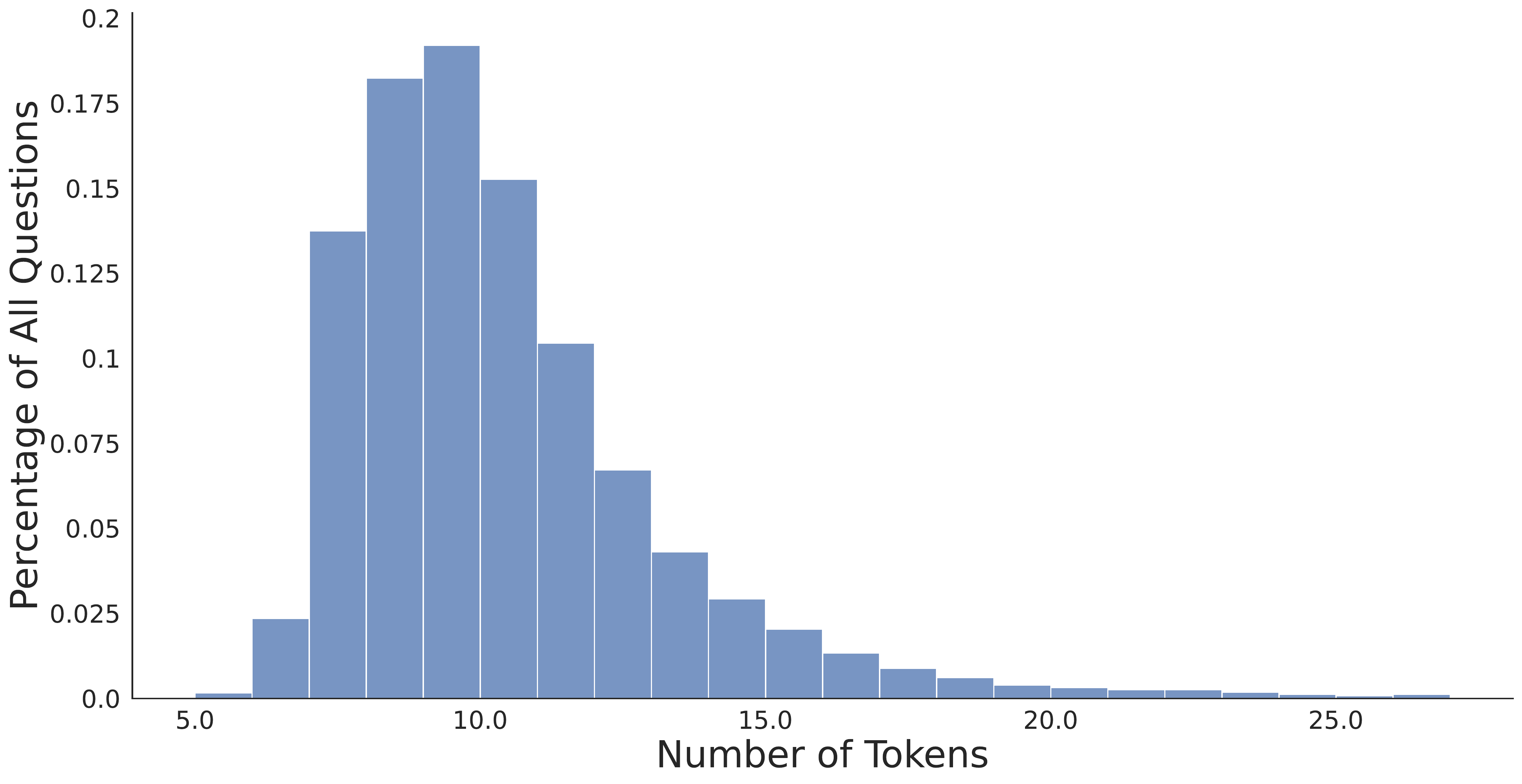}
    \caption{Sentence Length Distribution.}\label{fig:q_len_dist}
    % \vspace{-1pc}
    \label{fig:q_len_dist}
\end{figure}

Table~\ref{tb:dataset_statistics} reports the statistics of the crowdsourcing
task. Figure~\ref{fig:q_len_dist} shows the histogram of the sentence length of
the questions in \dataset, where the average question length is 10 tokens (Std=3).
Table~\ref{tb:top_frequent_ngram} list the top-15 frequent n-gram (with n=3) of questions opening in \dataset; this suggests that users on social media tend to
ask open-ended questions (beginning with ``Have you ever'', ``Do you think'', or
``How do you''), inviting others to share their opinions and expand the
conversation. The top-30 frequent words in \dataset are listed in
Table~\ref{tb:top_frquent_words}; this demonstrates that the
questions we gathered contain subjective words such as \textit{like},
\textit{think}, \textit{favorite}, and \textit{feel}, indicating that the
collected questions are more related to people's perspectives than objective
facts, encouraging individuals to answer them.

\subsection{Disentangling \dataset Effectiveness}

\begin{table}[t]
    \centering
    \small
    \begin{tabular}{lll}
        \toprule
        Have you ever & What would you & When was the\\
        Do you like & What do you & If you could\\
        Do you think & What kind of & Can you share\\
        What is the & How do you & Which is your\\
        What is your & Why do these & Does anyone know\\
        \bottomrule
    \end{tabular}
    % \vspace{-0.5pc}
    \caption{Top 15 frequent 3-gram of questions opening in \dataset.}
    % \vspace{-1.5pc}
    \label{tb:top_frequent_ngram}
\end{table}

\begin{table}[t]
    \centering
    \small
    \begin{tabular}{llllll}
        \toprule
        like & last & time & event & city & man\\
        people & party & place & wedding & know & enjoy\\
        ever & favorite & see & type & food & day\\
        would & go & kind & anyone & get & play\\
        think & many & family & friends & feel & best\\
        \bottomrule
    \end{tabular}
    % \vspace{-0.5pc}
    \caption{Top 30 frequent words in \dataset.}
    % \vspace{-1.5pc}
    \label{tb:top_frquent_words}
\end{table}

\begin{table*}[t]
    \centering
    \small
    \begin{tabular}{lccccc}
        \toprule
        Dataset & Vocab. Size $\uparrow$ & Avg. Sentence Length $\uparrow$ & Yngve Score $\uparrow$ & \% of Abstract Terms $\uparrow$ & Avg. Term Depth $\downarrow$ \\
        \midrule
        \midrule
        $VIST_1$ & 568 & 11.036 & 2.206 & \textbf{0.131} & 7.832\\
        $VIST_5$ & 592 & 11.165 & 2.173 & 0.127 & \textbf{7.406}\\
        $VQG_{orig}$ & 360 & 6.882 & 1.831 & 0.106 & 7.837\\
        $VQG_{ours}$ & \textbf{608} & \textbf{12.341} & \textbf{2.271} & 0.127 & 7.906 \\
        \bottomrule
    \end{tabular}
    % \vspace{-0.5pc}
    \caption{Comparison of question quality among different setups. $\uparrow$ indicates higher is better, $\downarrow$ indicates lower is better. The best scores are set in \textbf{bold}.}
    % \vspace{-1.5pc}
    \label{tb:dataset_analysis}
\end{table*}

\paragraph{Experimental Settings}

Two sources contribute to the efficacy of \dataset questions: 
1)~our question annotation approach, and 2)~the cohesive narrative of events
resulting from the five-photo arrangement. We conducted an experiment to
investigate the effect of these two factors on the question quality.

First, to evaluate the influence of the annotation approach, we selected VQG images
and annotated them with different instructions. We randomly chose 200 samples
from VQG and hired one worker per sample to annotate the image with our
instruction. The annotated questions ($\mathit{VQG}_{\mathit{ours}}$) were then compared to the
questions collected with original VQG instruction ($\mathit{VQG}_{\mathit{orig}}$).
Then, to evaluate the effect of the number of images, we randomly selected 200
samples from \dataset. For each five-image sample, we randomly
chose one image and hired one worker to annotate the selected image
per our instructions. The questions with the one-photo setup ($\mathit{VIST}_1$) were then
compared to the questions with the original five-photo \dataset setup
($\mathit{VIST}_5$). 

\paragraph{Quality Criteria}

Following \citet{ferraro-etal-2015-survey}, we evaluated the quality of
questions according to the following criteria:

\begin{itemize}[nosep,leftmargin=*,itemsep=0mm]
    \item Vocabulary size: the number of unique vocabulary words.
    \item Average sentence length: this shows how rich and descriptive the sentences are~\citep{ferraro-etal-2015-survey}. 
    % \mh{
    Writing a sentence is a high-cognitive task. However, to complete numerous jobs fast, MTurk workers typically write short and simple sentences (e.g., “What is the girl doing?”). These short questions are not detailed and are frequently similar to those from other workers. In other words, long question requires more effort from MTurk workers and can be more diverse, which may lead to higher quality.
    % }
	 \item Syntactic complexity: the amount of embedding/branching in a
	 sentence's syntax. We report the mean Yngve score~\citep{10.2307/985230}
	 normalized by the sentence length.
	 \item Percentage of abstract terms: this indicates the range of visual and
	 non-visual concepts covered by the dataset. 
    % Abstract terms are ideas or concepts, such as ``love'' or ``think;'' concrete terms are all the objects or occurrences that are primarily available to the senses, such as ``car'' or ``dog.'' 
    % We searched all noun tokens on WordNet~\citep{wordnet}, tokens belonging to \textit{Abstract Entity} are regarded as abstract terms, while tokens belonging to \textit{Physical Entity} are considered concrete terms. 
	 Of all noun tokens on WordNet~\citep{wordnet}, tokens belonging to
	 \textit{Abstract (Physical) Entity} are regarded as abstract (concrete)
	 terms.
	 \item Average term depth: noun terms on WordNet with a smaller depth
	 indicate higher-level concepts~\citep{liu-etal-2021-visually}. 
\end{itemize}
    
\paragraph{Results}

The first two columns in Table~\ref{tb:dataset_analysis} show that questions in $\mathit{VQG}_{\mathit{ours}}$
have a 1.7 times larger vocabulary size and are about 2 times longer on average 
than questions in $\mathit{VQG}_{\mathit{orig}}$, which reflects the fact that the proposed annotation
approach yields more diverse and descriptive sentences.
The third and forth columns in Table~\ref{tb:dataset_analysis} indicate that questions in $\mathit{VQG}_{\mathit{ours}}$
exhibit more complex sentence structure and have more abstract words than $\mathit{VQG}_{\mathit{orig}}$,
implying that writing down descriptions first helps individuals think more
about the abstract events behind the images and thus yields more complex
questions. This then makes our collected questions much easier for individuals to engage with.
The last column in Table~\ref{tb:dataset_analysis} shows that questions in $\mathit{VIST}_5$ have a smaller
term depth than questions in $\mathit{VIST}_1$, suggesting that questions in $\mathit{VIST}_5$
use more high-level concepts. \textit{Basic-level categories} were typically
used to name things~\citep{ROSCH1975573, Anglin1977WordOA, Brown1958HowSA},
whereas in multi-image scenarios, higher-level ideas were more often used to
cover things in various photos~\cite{categories_and_concepts}. This 
encourages individuals to answer questions not only based on the things they
saw, but by imagining the story or the relations of objects in the five
images.
This shows that our instructions contributed more to the engagement of the
collected questions than the multi-image setting. 

\section{Baselines}

We propose both end-to-end and dual-stage VL baselines for \dataset. We
introduce each baseline here and provide the details in 
Appendix~\ref{ap:baselines_implementation}.

\begin{table*}[t]
    \centering
    \small
    \begin{tabular}{clccccc}
        \toprule
        Group & Baseline & Benchmark~1 & Benchmark~2 & Benchmark~3 & Benchmark~4 & Benchmark~5 \\
        \midrule
        \midrule
        
        \multirow{3}{*}{1} & $\text{VL-T5}_\text{F\_VIST}$ & \textbf{35.08\%} (\textbf{1.981}) & \textbf{34.80\%} (\textbf{1.985}) & 34.60\% (1.983) & \textbf{35.04\%} (\textbf{1.971}) & \textbf{35.80\%} (\textbf{1.963}) \\
        & $\text{VL-T5}_\text{F\_VCR}$ & 31.84\% (2.038) & 32.64\% (2.018) & 30.64\% (2.050) & 32.60\% (2.026) & 30.12\% (2.058) \\
        & $\text{VL-T5}_\text{F\_VQG}$ & 33.08\% (1.981) & 32.56\% (1.995) & \textbf{34.76\%} (\textbf{1.966}) & 32.36\% (2.001) & 34.08\% (1.978) \\
        \midrule
        
        \multirow{3}{*}{2} & $\text{CAP2Q}_\text{CLIP}$ & 33.44\% (2.016) & \textbf{33.96\%} (2.019) & 32.88\% (2.020) & \textbf{34.24\%} (\textbf{1.986}) & 33.20\% (2.001) \\
        & $\text{STY2Q}_\text{CLIP}$ & \textbf{33.44\%} (2.006) & 32.80\% (1.995) & 33.36\% (2.001) & 31.64\% (2.025) & 33.32\% (2.002) \\
        & $\text{SUM2Q}_\text{CLIP}$ & 33.12\% (\textbf{1.977}) & 33.24\% (\textbf{1.985}) & \textbf{33.76\%} (\textbf{1.979}) & 34.12\% (1.988) & \textbf{33.48\%} (\textbf{1.996}) \\
        \midrule
        
        \multirow{3}{*}{3} & $\text{STY2Q}_\text{CLIP}$ & 34.09\% (1.988) & \textbf{34.81\%} (\textbf{1.974}) & \textbf{34.13\%} (\textbf{1.984}) & 32.01\% (2.022) & \textbf{33.97\%} (\textbf{1.993}) \\
        & $\text{VL-T5}_\text{F\_VIST}$ & \textbf{34.25\%} (\textbf{1.976}) & 32.97\% (2.012) & 32.97\% (2.008) & 33.89\% (\textbf{1.980}) & 32.13\% (2.008) \\
        & $\text{VL-T5}_\text{C}$ & 31.65\% (2.036) & 32.21\% (2.012) & 32.89\% (2.008) & \textbf{34.09\%} (1.998) & 33.89\% (1.999) \\
        \midrule
        
        \multirow{2}{*}{4} & $\text{VL-T5}_\text{F\_VIST}$ & \textbf{50.68\%} (\textbf{1.493}) & \textbf{51.24\%} (\textbf{1.488}) & \textbf{50.48\%} (\textbf{1.495}) & \textbf{50.48\%} (\textbf{1.495}) & 49.28\% (1.507) \\
        & $\text{VL-T5}_\text{A\_VIST}$ & 49.32\% (1.507) & 48.76\% (1.512) & 49.52\% (1.505) & 49.52\% (1.505) & \textbf{50.72\%} (\textbf{1.493}) \\
        \bottomrule
    \end{tabular}
    % \vspace{-.5pc}
    \caption{Human evaluation of different groups for five benchmarks. 
    Group~1: end-to-end baselines pretrained on different datasets. 
    Group~2: dual-stage baselines with different types of text as input. 
    Group~3: baselines with or without story information.
    Group~4: end-to-end baselines with fine-tuning or adapt-tuning. 
    Given methods and benchmarks by row and column, the percentage indicates the ratio of rank-1 questions among all questions (higher is better). The number in brackets is the average ranking among all questions (lower is better).}
    % \vspace{-.5pc}
    \label{tb:human_eval}
\end{table*}

For the end-to-end baselines, we chose the VL-T5 model~\citep{cho2021vlt5} as
the backbone
% because it handles all VL tasks as text-generating tasks, which is appropriate for our question generation scenario
. VL-T5 inputs contain the
visual embedding~$\mathcal{V}$ and the visual semantic grounding~$\mathcal{G}$.
Each image $V_i$ is handled as a sequence of visual embeddings
% $V_i=\{v^i_0, v^i_1, ..., v^i_k\}$ 
consisting of the whole image embedding
% $v^i_0$ 
and its 
% $k$ 
object region embeddings.
% $v^i_1$ to $v^i_k$
% Each visual embedding 
% $v^i_j$ 
% includes (1)~RoI (region of interest) features
% : the bounding box’s hidden representation generated by a ResNet50~\cite{DBLP:journals/corr/HeZRS15} model
% , (2)~RoI bounding box coordinates
% : the upper left and the lower right points of the box and its area
% , (3)~image positional indices
% : $\iota_{\mathit{img}} \in \{1,...,n\}$, where $n$ is the number of images, used to discriminate regions from different images
% , and (4)~object positional indices
% : $\iota_{obj} \in \{0,...,k\}$, which serve as the positional embeddings in an image
% .
% ; these are all projected to 768-dimensional vectors, summed, and layer-normalized to generate the final visual embedding $v^i_j$. Note that $\iota_{obj}$ in $v^i_0$ is~0.
As visual embeddings from RoI features lack semantic meaning, we inject visual
semantic grounding into VL-T5 to facilitate semantic understanding and
cross-image reasoning. We adopt grounded situation recognition
(GSR)~\cite{Pratt2020Swig} and the corresponding JSL model to produce
structured semantic summaries of images. For each image $V_i$, JSL outputs a
verb representing the salient activity of $V_i$ and its 1 to 6 corresponding
semantic roles. The predicted verb and nouns are combined as the visual
semantic grounding~$G_i$ of each image. 
% In particular, when tokenizing, we quote the verb with the starting and ending tokens \texttt{<b\_verb>} and \texttt{<e\_verb>} to highlight the activity, and the \texttt{<b\_[role]>} and \texttt{<e\_[role]>} tokens to spot the roles and their types. 
% The embeddings of text tokens for these semantic roles are randomly initiated during training, and each text embedding is combined with the same image's positional index embedding of its associated visual embedding $V_i$ to link the semantic role tokens to their corresponding visual images.

We propose three fine-tuned versions of VL-T5 respectively
pretrained on VCR~\citep{zellers2019vcr} ($\textbf{VL-T5}_\textbf{F\_VCR}$),
VIST ($\textbf{VL-T5}_\textbf{F\_VIST}$), and VQG
($\textbf{VL-T5}_\textbf{F\_VQG}$), and fine-tuned on \dataset. 
% We used the AdamW optimizer with a learning rate of 1e-4 and a batch size of 8 for both pretrained and fine-tuned tasks. When inference, we used nucleus sampling with $p = 0.9$, which has been shown to be effective in generating diverse text~\cite{DBLP:journals/corr/abs-1904-09751}.
\citet{DBLP:journals/corr/abs-1910-07117} show that after standard
fine-tuning, the model forgets important language generation skills acquired
during pretraining. Therefore, we propose the adapt-tuned version of VL-T5 by
adding the adapter layer to each Transformer block of the baseline, and
replacing the fine-tuning stage with adapt-tuning. In the adapt-tuning
stage, we update only the parameters of the adapter layer and freeze all other
parameters. We pretrain our model on VIST ($\textbf{VL-T5}_\textbf{A\_VIST}$)
and VQG ($\textbf{VL-T5}_\textbf{A\_VQG}$), and adapt-tune on \dataset. 
Moreover, inspired by \citet{shen2022how}, which shows that the CLIP visual
encoder~\citep{Radford2021LearningTV} can be used as visual embedding and
improve the VL model performance, we propose the CLIP version of VL-T5 by
replacing the visual embeddings of VL-T5 with the output of the CLIP visual
encoder ($\textbf{VL-T5}_\textbf{C}$). 
% We also create a baseline by removing the GSR information from $\textbf{VL-T5}_\textbf{C}$ ($\textbf{VL-T5}_\textbf{C-GSR}$).
% The CLIP variant we used is CLIP-RN50 (ResNet50~\cite{DBLP:journals/corr/HeZRS15} as visual backbone).

For the dual-stage baselines, 
% we first proposed the \textbf{PRVQG} inspired by the VIST model created by~\citet{hsu-etal-2021-plot}. We plotted the question lines from a knowledge graph constructed by VIST and Visual Genome~\citep{10.1007/s11263-016-0981-7}, and generated the question with a Transformer-based decoder. We employed a sentence evaluator trained on human ranking data~\citep{hsu-etal-2022-learning}, which consist of stories and corresponding human ranking scores, to deliver a reward. We then multiplied the reward to the negative log-likelihood loss to optimize the generated questions toward human preference.
% The second dual-staged model is Description2Q.
we first used an image captioning model to generate a description from an image
sequence, after which we used a question generation model to generate a question from
the description. The image captioning model used was a VL-T5 model pretrained
on VCR, and the question generation model was a T5 model~\citep{2020t5}
pretrained on SQuAD~\citep{rajpurkar-etal-2016-squad}. We provided three types
of text as descriptions: (1)~captions from the VIST dataset (\textbf{CAP2Q}), 
(2)~stories from the VIST dataset (\textbf{STY2Q}), and (3)~summaries from Q2 in
\dataset (\textbf{SUM2Q}). The VL-T5 image captioning model and
question generation model were fine-tuned on these three description types,
respectively. 
% Inspired by \citet{shen2022how}, which shows that CLIP visual encoder~\citep{Radford2021LearningTV} can be used as visual embedding and improve the VL models' performance, 
As the end-to-end baselines used CLIP to encode visual input,
we adopted the CLIP visual encoder in our dual-staged baselines. We replaced the
T5 model in the second stage with VL-T5 and then used the result
of the CLIP visual encoder as visual input and the descriptions as textual input.
For the different types of descriptions, we propose $\textbf{CAP2Q}_\textbf{CLIP}$,
$\textbf{STY2Q}_\textbf{CLIP}$, and $\textbf{SUM2Q}_\textbf{CLIP}$. 
% The details of our baselines' implementation are in Appendix~\ref{ap:baselines_implementation}.
% The CLIP variant we used is CLIP-RN50 (ResNet50 as visual backbone).

\section{Experiment and Discussion}

We randomly divided \dataset into the training (70\%), val (20\%), and test (10\%)
sets, and evaluated the models introduced earlier with human and automatic
metric evaluation.

\begin{table*}[t]
    \centering
    \small
    \begin{tabular}{lccccc}
        \toprule
        % \rowcolor{light_gray}
        \multicolumn{6}{c}{Ranking}\\
        \midrule
        Baseline & Benchmark~1 & Benchmark~2 & Benchmark~3 & Benchmark~4 & Benchmark~5\\
        \midrule
        $\text{VL-T5}_\text{F\_VIST}$& \textbf{35.1\%} & \textbf{34.9\%} & \textbf{34.0\%} & \textbf{34.9\%} & \textbf{34.0\%}\\
        $\text{VL-T5}_\text{F\_VCR}$ & 32.3\% & 31.7\% & 33.4\% & 32.4\% &33.4\%\\
        $\text{VL-T5}_\text{F\_VQG}$ & 32.5\% & 33.4\% & 32.6\% & 32.7\% &32.6\%\\
        \midrule
        \midrule
        % \rowcolor{light_gray}
        \multicolumn{6}{c}{Rating}\\
        \midrule
        Baseline & Benchmark~1 & Benchmark~2 & Benchmark~3 & Benchmark~4 & Benchmark~5\\
        \midrule
        $\text{VL-T5}_\text{F\_VIST}$& 3.55 & 3.68 & 3.66 & 3.66 & 3.69\\
        $\text{VL-T5}_\text{F\_VCR}$ & 3.60 & 3.59 & 3.62 & 3.61 & 3.61\\
        $\text{VL-T5}_\text{F\_VQG}$ & \textbf{3.66} & \textbf{3.68} & \textbf{3.72} & \textbf{3.72} & \textbf{3.70}\\
        \bottomrule
    \end{tabular}
    % \vspace{-.5pc}
    \caption{Evaluating end-to-end baselines pretrained on different datasets with human ranking and rating. For ranking, the percentage indicates the ratio of rank-1 questions among all questions. For rating, the score scaled from 1 (the worst) to 5 (the best). We set the highest scores in \textbf{bold}.}
    % \vspace{-.5pc}
    \label{tb:human_rating}
\end{table*}

\subsection{Human Evaluation}

Recent work has demonstrated the unreliability of automatic evaluation and
recommends relying on human evaluation~\cite{DBLP:journals/corr/LiuLSNCP16}.
Therefore, we first conducted a human evaluation to understand how people feel
about the generated questions, specifically whether they are natural,
engaging, and focus on high-level relations among objects. We
randomly selected 100 image sequences from the \dataset test set and generated
questions for each using our established baselines and models. For each
sequence, we hired five workers from Amazon MTurk to rank the generated
questions according to the following benchmarks:

\begin{itemize}[nosep,leftmargin=*,itemsep=0mm, label={}]
	 \item \textbf{Benchmark 1}: When you see images like these on social media,
	 it is natural to ask this question.
% 	 Natural questions are not about what is seen, but rather about what can be inferred given these objects~\cite{DBLP:journals/corr/MostafazadehMDZ16}. 
% 	 Given an image containing a boy opening presents, ``Is the gift for the boy’s birthday?''\ is natural, while ``What is the color of the gift?''\ is not.
	 \item \textbf{Benchmark 2}: This question focuses primarily on the
	 essential objects of the images and the relationships between these
	 objects.
% 	 Given an image sequence that describes a drunk man holding a glass at a party, ``Why does the man holding the cup look  so sleepy at the party?''\ is better than ``Why is the cup so small?''.
	 \item \textbf{Benchmark 3}: This question focuses primarily on the story or
	 event behind all the images rather than one specific image.
% 	 Given an image that contains a group of people dancing, and another image that contains a dog, ``Who brought the dog to the  party?''\ focuses on the story, whereas ``What is the dog's name?''\ does not.
	 \item \textbf{Benchmark 4}: This question is specific to the event where
	 the photos were taken. It could be irrelevant or weird to ask this question
	 for other similar events.
% 	 For example, ``Who is she?''\ is a generic question because it can appear in most circumstances, while ``Did the son receive good medical care?''\ is not.
	 \item \textbf{Benchmark 5}: This is an engaging question for this set of
	 photos. You would want to answer this question if you saw it on social
	 media.
    % A good question is one that fits within all of the four benchmarks listed above, or can also be evaluated by other criteria, such as fluency, logistics, or grammatical correctness.
\end{itemize}

Empirically, it is difficult for workers to rank many items at the same time; results
thereof are unreliable. Therefore, we divided our baselines into four
groups for further discussion. Results are shown in Table~\ref{tb:human_eval}.

\paragraph{Group 1: Different Pretrained Datasets}

First, we compare three VL-T5 baselines pretrained on VIST, VCR, and VQG,
respectively. The first group of results in Table~\ref{tb:human_eval} reveals that
the VL-T5 baseline pretrained on VIST performs best on most of the benchmarks.
The substantial difference between $\text{VL-T5}_\text{F\_VIST}$ and
$\text{VL-T5}_\text{F\_VCR}$ on Benchmark~5 suggests that story information
in the pretraining stage helps models ask more engaging questions.

\paragraph{Group 2: Image Description Type}

We compare three Description2Q baselines, each containing captions,
stories, and summaries as the input text. The result is displayed in the second
group of Table~\ref{tb:human_eval}. $\text{CAP2Q}_\text{CLIP}$ and
$\text{SUM2Q}_\text{CLIP}$ perform well on the Benchmarks~2 and~4 because
captions and summaries of photos are better able to provide details of objects
and lead to more specific questions. However, $\text{STY2Q}_\text{CLIP}$ has
the most rank-1 questions based on Benchmark~1. This suggests that story
information results in more natural questions. This finding also suggests
that naturalness may not be the main factor leading to engagement,
which contradicts the premise in VQG.  

\paragraph{Group 3: With or Without Story Information}

Third, we investigate the differences between baselines with and without story
information. We compare with-story baselines ($\text{VL-T5}_\text{F\_VIST}$ and
$\text{STY2Q}_\text{CLIP}$) and the without-story baseline
($\text{VL-T5}_\text{C}$). The result in the third group of
Table~\ref{tb:human_eval} shows that humans prefer questions generated by
baselines with story information. Moreover, the fact that $\text{STY2Q}_\text{CLIP}$
outperforms $\text{VL-T5}_\text{F\_VIST}$ on the Benchmark~5 suggests that the
generated questions could be even more engaging if the story information were
more explicit. 

\paragraph{Group 4: Fine-tuning and Adapt-tuning}

Finally, we compare the difference between fine-tuning and adapt-tuning
strategies on VL-T5 baselines pretrained on VIST. The result in the last group
of Table~\ref{tb:human_eval} shows that $\text{VL-T5}_\text{F\_VIST}$
outperforms $\text{VL-T5}_\text{A\_VIST}$ on Benchmarks~1 to~4, whereas
$\text{VL-T5}_\text{A\_VIST}$ surpasses $\text{VL-T5}_\text{F\_VIST}$ on 
Benchmark~5. Because adapt-tuning retains more information gained via VIST,
this result confirms the prior finding that explicit story information
results in engaging questions. 
Also note that this result shows that engagement
does not rely only on Benchmarks~1 to~4 as shown in related work.

\paragraph{Ranking vs. Rating}

% \mh{
In addition to ranking, several studies evaluated the generated text via human rating~\citep{Hu_Cheng_Gan_Liu_Gao_Neubig_2020, Wang_Wei_Li_Zhang_Huang_2020}. Though literature has shown that rating result is almost with no correlation with direct ranking~\citep{hsu2022learning}, here we still provide both results among $\text{VL-T5}_\text{F\_VIST}$, $\text{VL-T5}_\text{F\_VCR}$, and $\text{VL-T5}_\text{F\_VQG}$ for reference. For the rating experiment, we ask workers to rate the generated questions from 1 (the worst) to 5 (the best) according to the 5 benchmarks. We conduct both ranking and rating experiments on the whole testing set (N=599). The result in Table~\ref{tb:human_rating} shows that for ranking evaluation, $\text{VL-T5}_\text{F\_VIST}$ outperforms other two baselines on all benchmarks significantly (the Kruskal-Wallis test, \textit{p}=0.02), aligning the result in Table~\ref{tb:human_eval}, while for rating evaluation, $\text{VL-T5}_\text{F\_VQG}$ performs better insignificantly (\textit{p}=0.87). These results overall confirm that $\text{VL-T5}_\text{F\_VIST}$ is a better setting and \dataset should use ranking for evaluation.
%Since the two results are mismatched, we do the test of statistical significance on each result. For the ranking evaluation, we apply , a  statistical significance test for ranking data; for the rating evaluation, we apply t-test between $\text{VL-T5}_\text{F\_VIST}$ and $\text{VL-T5}_\text{F\_VQG}$. Results show that $\text{VL-T5}_\text{F\_VIST}$ is significantly better than $\text{VL-T5}_\text{F\_VCR}$ and $\text{VL-T5}_\text{F\_VQG}$ on ranking evaluation with p=0.02, whereas the rating results did not pass the significance test (p=0.87), which suggests that ranking is a better evaluation method for \dataset.
% }

\subsection{Automatic Evaluation}\label{sec:automatic_eval}

Although human evaluation is already a good indicator of model
performance, we still provide the automatic evaluation results here for
reference. We evaluate the baselines with BLEU~\citep{papineni-etal-2002-bleu},
METEOR~\citep{banerjee-lavie-2005-meteor}, and
BLEURT~\citep{sellam-etal-2020-bleurt}.
%, which are all reference-based metrics typically used to evaluate generation models. 
% In addition, to evaluate model diversity, we calculate the uniqueness (UNIQ) score, i.e., the ratio of inferred sentences that are unique within the generated sentences~\cite{park2020visualcomet}. 

%\paragraph{Results}

\begin{table}[t]
    \centering
    \small
    \begin{tabular}{lcccc}
        \toprule
        Baseline & B\_1 & B\_4 & METEOR & BLEURT\\
        \midrule
        \midrule
        $\text{VL-T5}_\text{F\_VCR}$ & 40.6 & 3.0 & 38.5 & -51.1\\
        $\text{VL-T5}_\text{F\_VIST}$ & \underline{42.7} & \textbf{4.8} & \textbf{41.8} & \textbf{-42.2} \\
        $\text{VL-T5}_\text{F\_VQG}$ & 41.3 & 3.6 & 40.1 & -46.6 \\
        $\text{VL-T5}_\text{A\_VIST}$ & 41.6 & 2.6 & 39.2 & \underline{-44.0} \\
        $\text{VL-T5}_\text{A\_VQG}$ & 41.2 & 3.2 & 38.9 & -51.1 \\
        $\text{VL-T5}_\text{C}$ & 41.6 & 3.8 & 38.7 & -54.0 \\
        % $\text{VL-T5}_\text{C-GSR}$ & 39.5 & 2.4 & 36.8 & -72.4 & 93.3\\
        % PRVQG & 32.9 & 1.9 & 29.1 & -70.0 & 37.5\\
        Cap2Q & \textbf{42.8} & 3.4 & 39.6 & -48.4 \\
        STY2Q & 41.1 & 3.4 & 39.6 & -48.6 \\
        SUM2Q & 41.7 & 3.0 & 39.5 & -47.4 \\
        $\text{CAP2Q}_\text{CLIP}$ & 40.6 & 3.3 & 40.5 & -46.4 \\
        $\text{STY2Q}_\text{CLIP}$ & 41.8 & \underline{4.2} & \underline{40.5} & -49.3 \\
        $\text{SUM2Q}_\text{CLIP}$ & 42.1 & 4.2 & 39.7 & -44.2 \\
        \bottomrule
    \end{tabular}
    % \vspace{-0.5pc}
    \caption{Automatic evaluation results of BLEU\_1~(B\_1), BLEU\_4~(B\_4), METEOR, and BLEURT. The highest (second-highest) scores are set in \textbf{bold} (\underline{underlined}).}
    % \vspace{-.5pc}
    \label{tb:autolatic_eval}
\end{table}

%The result of automatic evaluation is shown in 
Table~\ref{tb:autolatic_eval} shows the results. 
% \justin{what does the underlined results mean?}
$\text{VL-T5}_\text{F\_VIST}$ outperforms other baselines,
particularly $\text{VL-T5}_\text{F\_VCR}$ and $\text{VL-T5}_\text{F\_VQG}$. In
addition, $\text{STY2Q}_\text{CLIP}$ and $\text{SUM2Q}_\text{CLIP}$ outperform
other dual-stage baselines. These two results support the human evaluation
result: that models with story information generate more engaging questions.
Moreover, $\text{VL-T5}_\text{C}$ outperforming $\text{VL-T5}_\text{F\_VCR}$
and $\text{STY2Q}_\text{CLIP}$ outperforming STY2Q indicate that the CLIP
model provides better embeddings for question generation. 
% \mh{
Furthermore, all the dual-staged models with CLIP encoder outperform those without it. Since the second stage of those without CLIP generates questions from only text, and the second stage of those with CLIP generates questions from both texts and images, this result illustrates the assistance of visual information for \dataset.
% }
The only result
that differs from the human evaluation is that
$\text{VL-T5}_\text{F\_VIST}$ outperforms $\text{VL-T5}_\text{A\_VIST}$ and
$\text{STY2Q}_\text{CLIP}$. However, this is straightforward to explain: the
end-to-end fine-tuned model maintains the least information from pretraining
and leads to the most similar outcome to the fine-tuning data, which gives it an
advantage in the automatic metric evaluation where exact matches are rewarded.
% Moreover, the end-to-end baselines outperform two-staged baselines means that the end-to-end models avoid error propagation, which is a major drawback of two-staged models. $\text{VL-T5}_\text{F\_VIST}$ outperform $\text{VL-T5}_\text{A\_VIST}$ points out that although adapt-tuning keep more pretraining information, it may prevent models from learning the features of target tasks. In this case, the questions generated by $\text{VL-T5}_\text{A\_VIST}$ would be less engagement.

\subsection{Effect of Multi-Image Setting}\label{sec:ablation_study}

\begin{table}[t]
    \centering
    \small
    \begin{tabular}{lcccc}
        \toprule
        Baseline & B\_1 & B\_4 & METEOR & BLEURT\\
        \midrule
        \midrule
        
        $\text{VL-T5}_\text{F\_VIST}\ \spadesuit$ & \textbf{42.7} & 4.8 & \textbf{41.8} & \textbf{-42.2} \\
        $\text{VL-T5}_\text{F\_VIST}\ \diamondsuit$ & 42.1 & 3.9 & 39.5 & -50.5 \\
        $\text{VL-T5}_\text{F\_VIST}\ \clubsuit$ & 42.1 & \textbf{5.0} & 41.6 & -44.2 \\
        \midrule
        
        $\text{VL-T5}_\text{A\_VIST}\ \spadesuit$ & 41.6 & 2.6 & 39.2 & \textbf{-44.0} \\
        $\text{VL-T5}_\text{A\_VIST}\ \diamondsuit$ & \textbf{41.8} & \textbf{2.9} & 38.0 & -51.8 \\
        $\text{VL-T5}_\text{A\_VIST}\ \clubsuit$ & 41.4 & 2.7 & \textbf{39.8} & -44.2 \\
        \midrule
        
        $\text{STY2Q}\ \spadesuit$ & \textbf{41.1} & 3.4 & 39.6 & -48.6 \\
        $\text{STY2Q}\ \diamondsuit$ & 40.5 & 2.8 & 37.9 & -53.4 \\
        $\text{STY2Q}\ \clubsuit$ & 41.0 & \textbf{3.5} & \textbf{39.8} & \textbf{-48.4} \\
        \midrule
        
        $\text{STY2Q}_\text{CLIP}\ \spadesuit$ & \textbf{41.8} & \textbf{4.2} & \textbf{40.5} & \textbf{-49.3} \\
        $\text{STY2Q}_\text{CLIP}\ \diamondsuit$ & 41.3 & 2.9 & 38.0 & -53.4 \\
        $\text{STY2Q}_\text{CLIP}\ \clubsuit$ & 41.3 & 2.6 & 39.1 & -51.7 \\
        \bottomrule
    \end{tabular}
    % \vspace{-0.5pc}
    \caption{The effect of different input: $\spadesuit$ the whole image sequence, $\diamondsuit$ the most relevant image selected by CLIP score, and $\clubsuit$ the image sequence without the most relevant image. Results evaluated by BLEU\_1~(B\_1), BLEU\_4~(B\_4), METEOR, and BLEURT.}
    \label{tb:multi_image_comp}
    % \vspace{-.5pc}
\end{table}

We study the impact of the multi-image setting on beselines. Here we seek
to determine whether the most relevant image can represent the entire image sequence, as
questions can focus on only one certain event or object. 
We begin by
determining the most representative image in the image sequence by calculating
the CLIP score, the cosine similarity between each image and the ground truth
question. Then we examine questions generated from three types of input: 
(1)~the entire image sequence, (2)~only the most relevant image, and (3)~the image
sequence without the most relevant image. 
%We examine these three sorts of input on all baselines and evaluate them using the previously mentioned automatic metrics.

Table~\ref{tb:multi_image_comp} shows the experiment results. Using the most
relevant image leads to the lowest score in most of the baselines, implying
that a single image cannot in fact represent the whole image sequence and the underlying
event or scenario. Surprisingly, the results also show that even after removing the
most relevant image, 
the performance of some baselines is still high.    % AMH: check checked
This 
suggests that other images in the sequence assist in the reconstruction of
missing information and even leave room for more imagination. It also shows
that the collected questions %in Multi-VQG do 
cover information from all images
in the sequence.

\subsection{Case Study}

Table~\ref{tb:case} displays example image sequences and questions generated by baselines.
Cases~1 and~2 provide clues for the reason why human
evaluation and automatic metrics produce inconsistent results for
$\text{VL-T5}_\text{F\_VIST}$, $\text{VL-T5}_\text{A\_VIST}$, and
$\text{STY2Q}_\text{CLIP}$. In case~1, both the ground truth and the
$\text{VL-T5}_\text{F\_VIST}$ output mention the flower, whereas
$\text{VL-T5}_\text{A\_VIST}$ focuses on the insects and the bird. Because
fine-tuned models are more likely to forget the pretrained task and fit the
ground truth of the fine-tuned task, $\text{VL-T5}_\text{F\_VIST}$ may obtain a
higher score from the match-based automatic metrics. Adapt-tuning, on the other
hand, retains more information from the pretrained task and results in models that do not
always follow the guide of the ground truth. As a result, while
$\text{VL-T5}_\text{A\_VIST}$ may have a lower automatic evaluation score, it
may generate questions that follow the story, reflecting human preferences.
Case~2 shows the case of diverse images. As the first three photos are very
different from the last two, it is hard for
$\text{VL-T5}_\text{F\_VIST}$ to generate an engaging question using implicit
story information, resulting in a general question. $\text{STY2Q}_\text{CLIP}$,
in contrast, takes an explicit story as input, which enables the model to
generate a question connected to the underlying story.

\begin{table}[t]
    \centering
    \small
    \begin{tabular}{lp{0.65\linewidth}}
    \toprule
        \multicolumn{2}{c}{Case~1} \\
    \midrule
        \multicolumn{2}{c}{
        \includegraphics[width=0.45\textwidth]{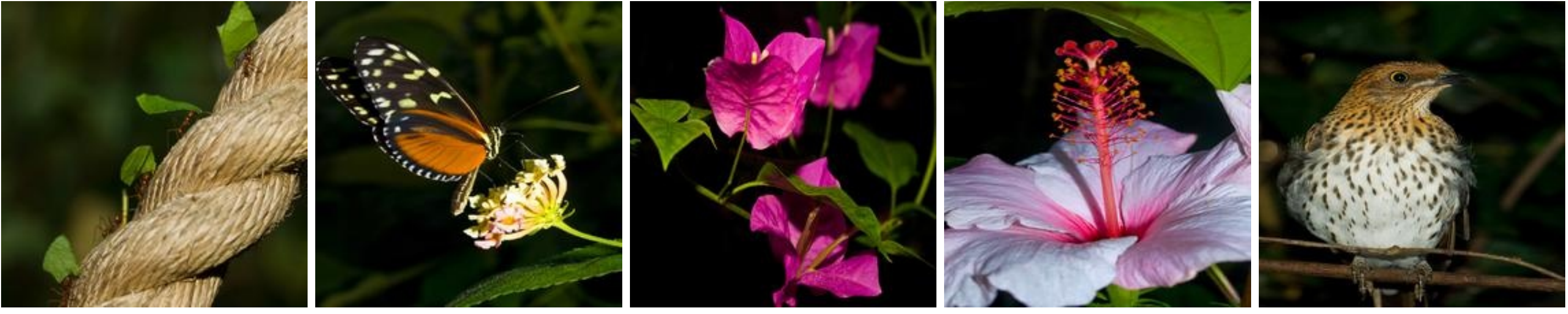}
        }\\
    \midrule 
        Ground Truth & Can someone tell me what the name of the bright pink flowers are?\\[1pt]
        \hdashline\noalign{\vskip 0.7ex}
        $\text{VL-T5}_\text{F\_VIST}$ & What color is the flower?\\[1pt]
        \hdashline\noalign{\vskip 0.7ex}
        $\text{VL-T5}_\text{A\_VIST}$ & Have you been interested in learning about bugs and bird life?\\
    \midrule 
    \midrule 
        \multicolumn{2}{c}{Case~2} \\
    \midrule
        \multicolumn{2}{c}{
        \includegraphics[width=0.45\textwidth]{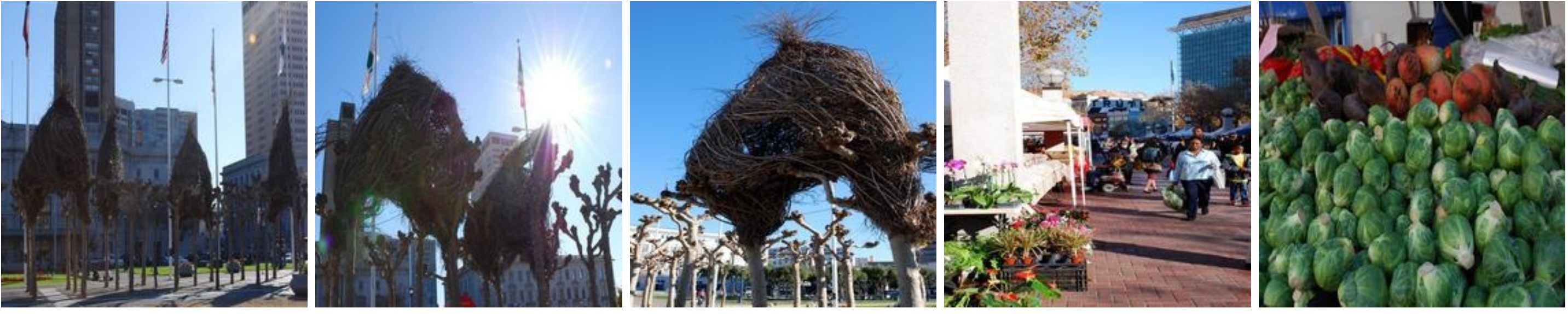}
        }\\
    \midrule 
        Ground Truth & When you look at these pictures, what else do you think might be sold?\\[1pt]
        \hdashline\noalign{\vskip 0.7ex}
        $\text{VL-T5}_\text{F\_VIST}$ & Do you like to go to places that have a crowd of people?\\[1pt]
        \hdashline\noalign{\vskip 0.7ex}
        $\text{STY2Q}_\text{CLIP}$ & What kind of food would you like to buy at this festival?\\
    \midrule 
    \midrule 
        \multicolumn{2}{c}{Case~3} \\
    \midrule
        \multicolumn{2}{c}{
        \includegraphics[width=0.45\textwidth]{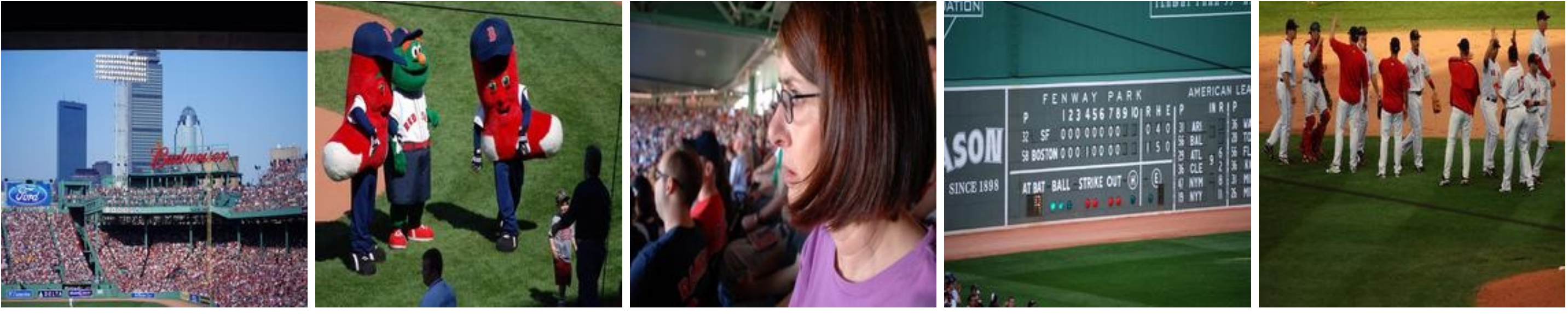}
        }\\
    \midrule 
        Ground Truth & How does everyone think the red sox are going to do this year?\\[1pt]
        \hdashline\noalign{\vskip 0.7ex}
        $\text{VL-T5}_\text{A\_VIST}$ & Are you fascinated by {\color{red}football}?\\[1pt]
        \hdashline\noalign{\vskip 0.7ex}
        $\text{STY2Q}_\text{CLIP}$ & Have you ever been to a {\color{red}costume party} before?\\
    \bottomrule
    \end{tabular}
    % \vspace{-0.5pc}
	 \caption{Questions generated by different methods.}
% 	 \vspace{-1.5pc}
    \label{tb:case}
\end{table}

Although $\text{STY2Q}_\text{CLIP}$ and $\text{VL-T5}_\text{A\_VIST}$ appear to
be better than $\text{VL-T5}_\text{F\_VIST}$, the generated questions may still
include errors. Case~3 is an example illustrating several commonly-seen
errors. First, an object and relationship detection error is observed in the
output of $\text{VL-T5}_\text{A\_VIST}$. The baseline mistakenly detects the
objects in the last image as football players. As a result, it asks ``Are you
fascinated by football?'' instead of ``baseball.'' Second, an inference error
is shown in the output of $\text{STY2Q}_\text{CLIP}$, where people are dressed
up in costumes in the second image but it mistakenly detects the event as a
costume party. Here we see that grounding and event inference are two major
directions for improving the quality of the generated questions.

\section{Conclusion}

We propose a novel task: given a sequence of images, 
generate an engaging question. This task extends visual question generation
by enabling reasoning across images to comprehend a complete story. We collect
\dataset by asking workers to write down five obvious objects, a summary of the
image sequence, as well as an engaging question that they would want to post on social
media. We establish several baselines for this task.
%, including both end-to-end
%and dual-stage models. 
Experimental results reveal that image-related stories
help models generate engaging questions, and that using multiple images as
input helps models understand the overall picture of the current situation,
leading to a better question. 
%However, there is still room to improve the quality of generated questions because of detecting wrong objects or relationships among entities and inferring incorrect events. 
The task, dataset, and experimental results we provide open up an exciting
challenge for visual-and-language models to implicitly construct a story behind
a series of photos for creativity and experience sharing and further
attracting attention for downstream applications.

\section*{Limitations}

Like most crowdsourced datasets, \dataset inherits the common biases of using online crowdsourcing platforms to collect data.
For example, the crowd workers on Amazon Mechanical Turk do not represent the user population of popular social media, such as Twitter.
Furthermore, although we instructed workers to write questions as if they were posting on Twitter, the used language would still be different.
People on social media use informal words and netspeak frequently, but crowd workers are incentivized to get their work approved and might prefer to use more formal languages or polite tones.
% \mh{
Moreover, since we specifically encouraged MTurk workers to imagine they are writing questions that they would ask on Twitter,~\dataset may be potentially biased on Tweet-liked data. 
%The reason for naming a specific social media platform is to reduce the variation in the collected data so that the generation model could learn better. 
We expect that different platforms will encourage different text styles, but given the amount of data we could financially afford to collect in the first study for this research problem, we decide to focus on only one platform’s style to reduce possible factors. 
%The reason for choosing Twitter is that it encourages shorter posts and is also popular worldwide. Admittedly, we could, instead, 
Asking workers to imagine Facebook or Instagram can be another practice, but it will still introduce different biases.
% }

Another limitation is the evaluation of engagement.
We evaluated the question engagement by asking crowd workers to rank the questions using different criteria. 
However, this approach does not capture the in-the-moment feelings or authentic reactions of social media users. 
The human evaluation results may not reflect the actual performance when the technology is being deployed in the wild.

\section*{Ethical Considerations}

% Kenneth edited
Although our research aims to produce natural and engaging questions, we are aware of the possibility of employing a similar approach to generate inappropriate,  sexist, or racist questions. 
Furthermore, as the proposed methods use a pre-trained grounded situation recognition and a T5 model as components, the generated questions might inherit the biases of their training data.
More research is required to understand and mitigate these risks.

\section*{Acknowledgements}

This research are partially supported by National Science and Technology Council,
Taiwan, under Grant no. 110-2634-F-002-051- and 108-2221-E-001-012-MY3.

\bibliographystyle{acl_natbib}
\bibliography{custom}

\clearpage
\appendix

\section{Implementation Details of Baselines}
\label{ap:baselines_implementation}

\subsection{VL-T5}\label{ap:VL-T5}

\begin{figure}[t]
     \centering
     \includegraphics[width=\linewidth]{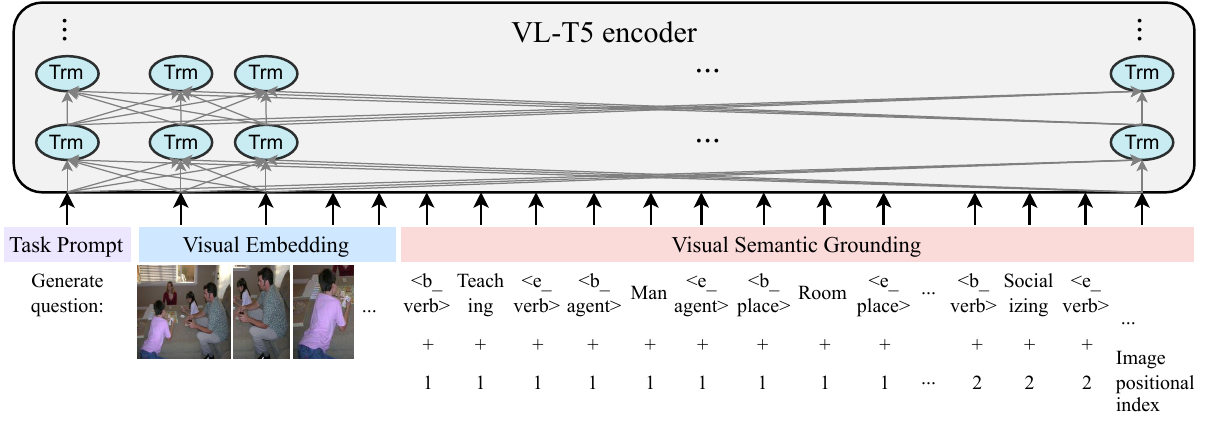}
% 	  \vspace{-1.8pc}
	  \caption{The input of VL-T5.
	  }
% 	  \vspace{-0.5pc}
\label{fig:vl-t5}
\end{figure}

The input of the VL-T5 model is depicted in Figure~\ref{fig:vl-t5}. The input
contains the task prompt, the visual embedding, and the visual semantic
grounding. Each semantic grounding embedding is the sum of the token embedding
and the image positional embedding. The semantic grounding is produced by
grounded situation recognition (GSR)~\cite{Pratt2020Swig} and the corresponding
JSL model. Consider the image in Figure~\ref{fig:vl-t5}, which depicts a man
teaching a boy. JSL predicts the primary activity \emph{teaching} (verb frame)
and then the agent \emph{man} and place \emph{room} as its semantic roles. The
predicted verb and nouns are combined as the visual semantic grounding~$G$ of
each image. In particular, when tokenizing, we quote the verb with the starting
and ending tokens \texttt{<b\_verb>} and \texttt{<e\_verb>} to highlight the
activity, and the \texttt{<b\_[role]>} and \texttt{<e\_[role]>} tokens to spot
the roles and their types, as illustrated in Figure~\ref{fig:vl-t5}.
The decoder, which is similar to the original T5 decoder, is omitted from
the figure for brevity. The embeddings of text tokens for these semantic roles
are randomly initiated during training, and each text embedding is combined
with the image's positional index embedding of its associated visual embedding
$V_i$ to link the semantic role tokens to their corresponding visual images.

\begin{figure}[t]
     \centering
     \includegraphics[width=0.85\linewidth]{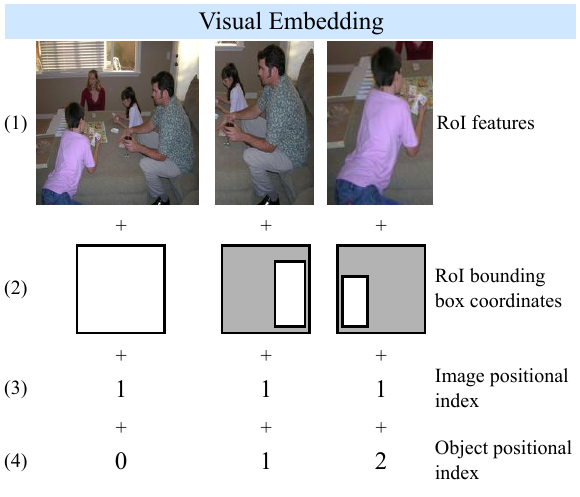}
% 	  \vspace{-1pc}
	  \caption{Details of visual embedding.
	  }
% 	  \vspace{-1.5pc}
\label{fig:visual_embedding}
\end{figure}

Figure~\ref{fig:visual_embedding} illustrates how images are encoded. Each
image $V_i$ is handled as a sequence of visual embeddings $V_i=\{v^i_0, v^i_1,
\dots, v^i_k\}$ consisting of the entire image embedding $v^i_0$ and its $k$
object region embeddings $v^i_1$ to $v^i_k$. Each visual embedding $v^i_j$
includes (1)~RoI features: the hidden representation of the bounding box
created by a ResNet50~\cite{DBLP:journals/corr/HeZRS15} model, (2)~RoI bounding
box coordinates: the upper left and the lower right points of the box and its
area, (3)~image positional indices: $\iota_{\mathit{img}} \in \{1,\dots,n\}$,
where $n$ is the number of images, used to discriminate regions from different
images, and (4)~object positional indices: $\iota_{obj} \in \{0,\dots,k\}$, which
serve as positional embeddings in an image. These are all projected to
768-dimensional vectors, summed, and layer-normalized to form the final visual
embedding $v^i_j$. Note that $\iota_{obj}$ in $v^i_0$ is~0. 

We used the AdamW optimizer with a learning rate of 1e-4 and a batch size of 8
for both pretrained and fine-tuned tasks. During inference, we used nucleus
sampling with $p = 0.9$, which has been shown effective in generating
diverse text~\cite{DBLP:journals/corr/abs-1904-09751}.

\begin{figure}[t]
     \centering
     \includegraphics[width=0.85\linewidth]{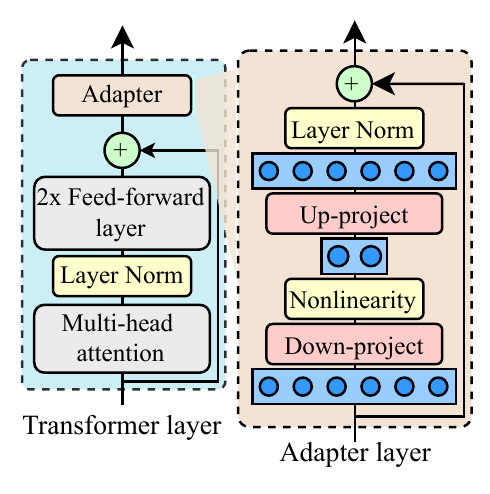}
% 	  \vspace{-1.2pc}
	  \caption{Details of the adapter layer.
	  }
% 	  \vspace{-1.5pc}
\label{fig:adapter}
\end{figure}

\subsection{Adapter Layer}

For the adapt-tuned baselines, we employed an adapter layer after the original
Transformer layers for both the VL-T5 encoder and decoder, as shown in
Figure~\ref{fig:adapter}. The adapter layer down-projects the input as a
384-dimensional vector, passing it into a GELU activation
function~\citep{https://doi.org/10.48550/arxiv.1606.08415}, up-projecting it to
the original size, and finally using a residual layer to sum the projected
vector with the input. During the pretraining stage, we bypassed the adapter
layers and trained only the parameters of the original Transformer layers. During
the adapt-tuning stage, we considered and trained the adapter layers while fixing
the parameters of the original parts.

\subsection{CLIP as Visual Encoder}

\begin{figure}[t]
     \centering
     \includegraphics[width=\linewidth]{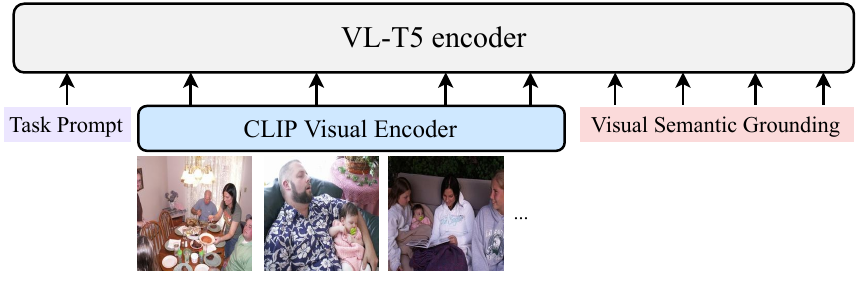}
% 	  \vspace{-1.8pc}
	  \caption{CLIP visual encoder in VL-T5.
	  }
% 	  \vspace{-0.5pc}
\label{fig:clip}
\end{figure}

Figure~\ref{fig:clip} depicts how the CLIP visual encoder is used in the VL-T5
baseline. Instead of finding RoI features and bounding boxes in each image, we
put the entire image into the CLIP visual encoder and obtained the visual
embedding. Because the visual embedding from CLIP was a 1024-dimensional vector,
we projected it onto 768 dimensions using a linear layer. The CLIP visual encoder
and the linear layer were tuned during the training stage. The CLIP variant we
used was CLIP-RN50 (ResNet50 as the visual backbone).

\subsection{Description2Q}

\begin{figure}[t]
     \centering
     \includegraphics[width=\linewidth]{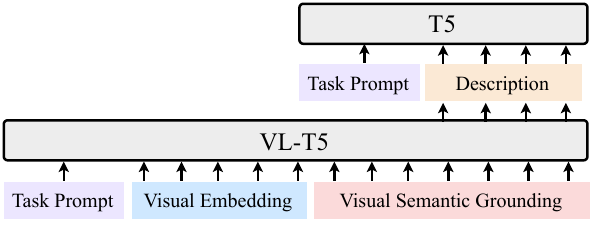}
% 	  \vspace{-1.8pc}
	  \caption{The architecture of Description2Q.
	  }
% 	  \vspace{-0.5pc}
\label{fig:description2q}
\end{figure}

The architecture of Description2Q is shown in Figure~\ref{fig:description2q}.
We used a VL-T5 model to generate descriptions from image sequences. The input
of VL-T5 was the same as in \ref{ap:VL-T5}, and the output description was
a caption, story, or summary, depending on the fine-tuning tasks. Then the
generated description was fed into a T5 model pretrained on
SQuAD and fine-tuned on the ground truth of descriptions to generate a
question.

% This is an appendix.

\end{document}